%% file: acl_latex.tex
\pdfoutput=1

\documentclass[11pt]{article}

\usepackage{acl}

\usepackage{times}
\usepackage{latexsym}
\usepackage{bm}
\usepackage{CJKutf8}

\usepackage[T1]{fontenc}

\usepackage[utf8]{inputenc}

\usepackage{microtype}

\usepackage{graphicx}
\usepackage{multirow}
\usepackage{amsmath}

\title{Building a Personalized Dialogue System with Prompt-Tuning}

\author{Tomohito Kasahara${}^\mathrm{1}$, Daisuke Kawahara${}^\mathrm{1}$, \\
\bf{Nguyen Tung${}^\mathrm{2}$, Shengzhe Li${}^\mathrm{2}$, Kenta Shinzato${}^\mathrm{2}$, Toshinori Sato${}^\mathrm{2}$}\\[3pt]
    ${}^\text{1}$Waseda University, ${}^\text{2}$LINE Corporation\\
    \small\texttt{\{tomo\_k@ruri.,dkw@\}waseda.jp} \\
    \small\texttt{\{tung.nguyen,shengzhe.li,kenta.shinzato,toshinori.sato\}@linecorp.com} \\}

\begin{document}
\maketitle
\begin{abstract}
Dialogue systems without consistent responses are not fascinating. In this study, we build a dialogue system that can respond based on a given character setting (persona) to bring consistency. Considering the trend of the rapidly increasing scale of language models, we propose an approach that uses prompt-tuning, which has low learning costs, on pre-trained large-scale language models. The results of automatic and manual evaluations in English and Japanese show that it is possible to build a dialogue system with more natural and personalized responses using less computational resources than fine-tuning.
\end{abstract}

\input{intro}
\input{relatedwork}
\input{method}
\input{experiments}
\input{conclusion}

\input{acknowledgements}

\bibliography{anthology,custom}
\bibliographystyle{acl_natbib}

\clearpage
\input{appendix}

\end{document}

%% file: intro.tex
\section{Introduction}
\label{sec:intro}
Large dialogue corpora used to train dialogue systems using neural network models contain utterances from various speakers.
This has the disadvantage that the trained system is often inconsistent in the generated utterances~\citep{li-etal-2016-persona}. For example, after the system says, ``I am from Tokyo,'' it might say, ``I am from Kyoto.''

We aim to build a dialogue system that can respond based on a persona to avoid inconsistent utterances.
A simple method of giving a persona to a model can be to concatenate the persona to the model's input in natural language~\citep{zhang-etal-2018-personalizing}. However, this method is not suitable because the more persona information is added, the longer the input text becomes.
Therefore, we propose to freeze all parameters of a pre-trained language model and add a new fixed-length prompt before the input token sequence to embed the persona information.
Specifically, only the embedding vectors of the added prompt are optimized using a dialogue corpus in which utterances are made based on the persona.

We conduct experiments on two languages: English and Japanese. Automatic and manual evaluations show that our method can build a dialogue system capable of natural responses based on a persona. Since our approach does not update the parameters of the pre-trained model, it can reduce the computational cost required for training.
We also show that it is possible to build a personalized dialogue system with even a small dataset consisting of hundreds to thousands of utterance-response pairs.

%% file: relatedwork.tex
\section{Related Work}
\label{sec:relatedwork}

\subsection{Prompt-Tuning}
With the advent of pre-trained models such as BERT~\citep{devlin-etal-2019-bert} and T5~\citep{JMLR:v21:20-074}, a method that adapts a pre-trained model to a target task by fine-tuning has become mainstream.
However, as the scale of models grows and the cost of fine-tuning increases, methods for adapting a pre-trained model to a target task without updating their parameters are gaining attention.

~\citet{NEURIPS2020_1457c0d6} proposed a zero/few-shot learning method based on language models with manually created task descriptions and zero/a few task examples (collectively called \textit{prompt}).
Although there are some studies on improving this method~\citep{reynolds2021prompt, pmlr-v139-zhao21c}, they are inferior to fine-tuning in terms of accuracy.

Prompt-tuning is a method for automatically optimizing a prompt without creating it by manual labor. 
There are two kinds of methods in prompt-tuning: one is to select the best words from a discrete vocabulary~\citep{shin-etal-2020-autoprompt}, and the other is to optimize continuous embedding vectors~\citep{qin-eisner-2021-learning, li-liang-2021-prefix, lester-etal-2021-power, DBLP:journals/corr/abs-2103-10385, DBLP:journals/corr/abs-2110-07904}.
Prefix-tuning~\citep{lester-etal-2021-power, li-liang-2021-prefix} adds a sequence of tokens, called prefix tokens, to the beginning of the input and optimizes only their embedding vectors.
There is also a study on multimodal prompt-tuning for images and natural language~\citep{tsimpoukelli2021multimodal}.


\subsection{Persona-Based Dialogue Systems}
According to~\citet{roller-etal-2021-recipes}, for dialogue systems to interact more naturally with humans, it is essential to consider three perspectives: having a consistent personality, having knowledge, and having emotions and empathy for the interlocutor. Among these three perspectives, we focus on personality because we believe that it is the most important to generate consistent responses.

The Persona-Chat dataset~\citep{zhang-etal-2018-personalizing} is a dataset created with the goal of adding personality to a dialogue system. It consists of multi-turn dialogues between two crowdworkers, each of whom is given approximately five persona sentences, which describe their character settings.
There are 1,155 personas in the Persona-Chat dataset.
There are two types of persona sentences per persona: \textit{original}, which the worker used in the dialogue, and \textit{revised}, which is a paraphrased version of the original.
In the experiments conducted by \citet{zhang-etal-2018-personalizing}, models were trained using all the data in the Persona-Chat dataset, which contains utterances based on various personas. On the other hand, our method uses dialogue data uttered based on only one persona to train models.
There is also a Japanese version of the Persona-Chat dataset, JPersonaChat~\citep{sugiyama2021empirical}.
Other dialogue corpora that contain speaker persona information include PersonalDialog~\citep{DBLP:journals/corr/abs-1901-09672} and a corpus of dialogue data from Reddit~\citep{mazare-etal-2018-training}.
~\citet{DBLP:journals/corr/abs-1901-09672} proposed a method to add encoded persona information to the input before it is fed into a seq2seq model.

%% file: method.tex
\section{Method}
\label{sec:method}
This section describes our proposed method. The detailed setup for our experiments is described in Sections~\ref{ssec:dataset_creation} and \ref{ssec:model_setup}.

\subsection{Proposed Model}
\begin{figure}[t]
    \centering
    \includegraphics[width=1\linewidth]{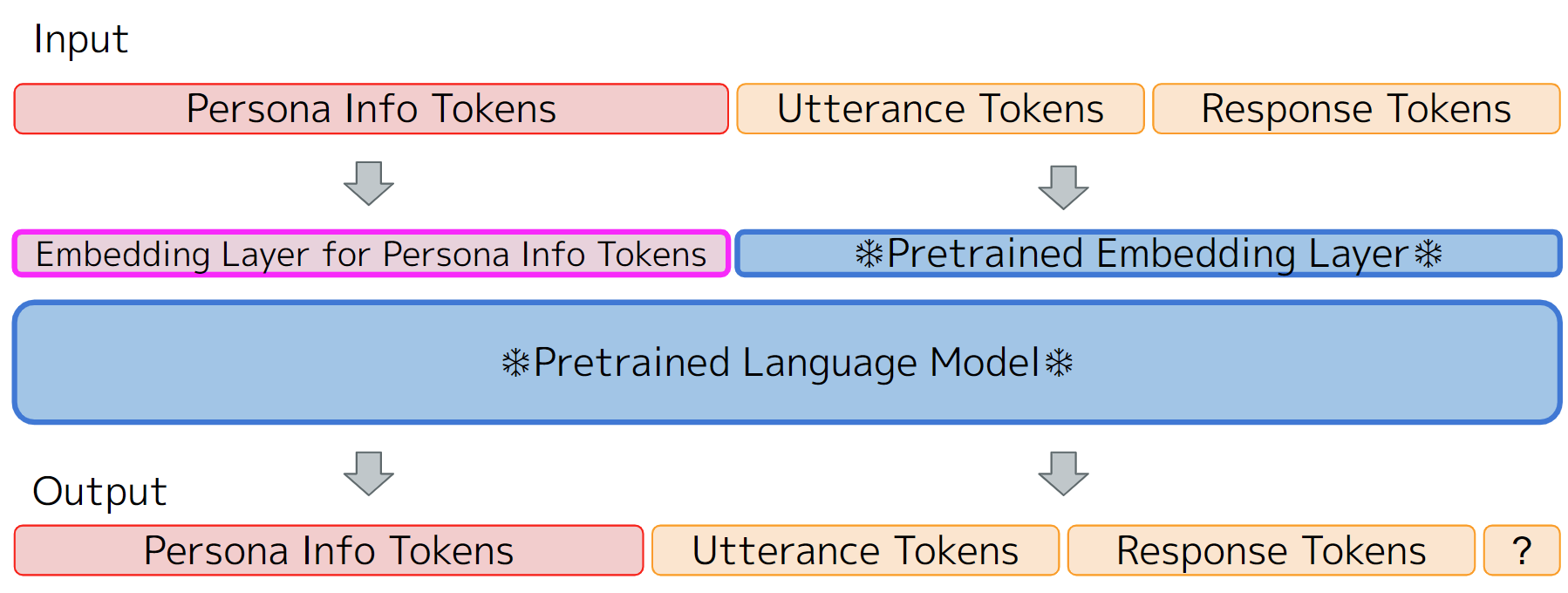}
    \caption{Architecture and input-output relation of the proposed model. All parameters of the pre-trained language model and its embedding layer are frozen. Only the newly added embedding layer for persona info tokens is tuned.}
    \label{fig:model}
\end{figure}

We propose a Transformer-based model with an additional embedding layer for tokens that embed persona information. We refer to these tokens as \textit{persona info tokens}.
The architecture and input-output relation of the proposed model are shown in Figure~\ref{fig:model}.

\subsection{Datasets}
Conversations in daily life are not always related to personal information~\citep{song-etal-2021-bob}. To allow the model to generate not only utterances that are related to the persona but also utterances that are not related to the persona, we make a dialogue dataset that consists of two types of datasets.
The first is a dialogue dataset where each utterance is based on the persona, and the second is a dialogue dataset that is not related to the persona.

\subsection{Training}
The newly added embedding layer embeds persona info tokens, and the embedding layer of the pre-trained language model embeds each pair of utterance and response (which consists of tokens already generated during training).
These embedding vectors are combined and then input into the model.
During training, the cross-entropy loss is calculated for the output tokens of the response sentence, and only the parameters of the embedding layer for the persona info tokens are updated.

The embedding layer for the persona info tokens is initialized with the persona sentences included in the Persona-Chat dataset. These sentences are embedded into vectors by the embedding layer of the pre-trained language model and then used for initialization. If the number of the tokens of the persona sentences is less than the length of the persona info tokens, the persona sentences are repeatedly arranged until the number is satisfied.

%% file: experiments.tex
\section{Experiments}
\label{sec:experiments}
Based on the method in Section~\ref{sec:method}, we build a personalized dialogue system.
We used Hugging Face's Transformers to build the system and the NVIDIA A100 SXM4 GPU with a GPU memory size of 40 GB.
The main experiments are conducted in English, and the results of additional experiments in Japanese are included at the end of this section.

\subsection{Datasets Setup}
\label{ssec:dataset_creation}
We use the Persona-Chat dataset\footnote{\url{https://github.com/facebookresearch/ParlAI/tree/main/parlai/tasks/personachat}} and DailyDialog~\citep{li-etal-2017-dailydialog}\footnote{\url{https://aclanthology.org/I17-1099/}} for our experiments in English.

\subsubsection{Training Datasets}
\label{ssec:training_dataset}
First, the multi-turn dialogues in the Persona-Chat dataset are divided into two utterances of one round trip. We refer to this pair of two utterances as \textit{a dialogue pair}. The dialogue pairs are aggregated according to the persona type given to the responder.
There are 1,155 personas in the Persona-Chat dataset, but we use the three personas with the most dialogue pairs in our experiments.
The reason for this is that we intend to experiment with a relatively large number of dialogue pairs even in the small dataset.
The number of dialogue pairs based on these three personas is 185, 167, and 166, respectively.
Three models corresponding to the three personas are trained and evaluated for each experimental setup.
The aggregated dialogue pairs are divided into training and evaluation pairs in a ratio of 9:1.

The Persona-Chat dataset does not contain many short utterances or utterances unrelated to persona.
To add utterances that are short and not related to persona to the dataset, 
we also use dialogue pairs contained in DailyDialog whose topic is Relationship,\footnote{Each dialogue is assigned a topic. There are ten topics: Attitude \& Emotion, Culture \& Education, Finance, Health, Ordinary Life, Politics, Relationship, School Life, Tourism, and Work.} which contains many such utterances.
Among them, dialogue pairs in which the lengths of both the utterance and the response are less than 50 characters are mixed into the training datasets in a certain ratio.
Based on the results of preliminary experiments, we determined the ratio of dialogue pairs added from DailyDialog to the number of those obtained from the Persona-Chat dataset as 1:1. We call this \textit{the ratio of the training datasets}.

\subsubsection{Evaluation Datasets}
We made two datasets for evaluation: the persona eval dataset and the general eval dataset. The persona eval dataset is 10\% of the 9:1 dataset described in Section~\ref{ssec:training_dataset}. The general eval dataset consists of dialogue pairs obtained from DailyDialog under the same conditions as in Section~\ref{ssec:training_dataset}, but not used for training.

        
        

\begin{table}[t]
\centering
\small
\begin{tabular}{c|c||c|c}
\hline
\textbf{Training Method} & \textbf{Model} & \textbf{Dist-1} & \textbf{Dist-2}\\\hline
\hline 

Fine-Tuning (added) & \multirow{3}{*}{GPT2-XL} & 0.199 & 0.526\\\cline{1-1}
Fine-Tuning (none) &  & 0.210 & 0.568\\\cline{1-1}
\multirow{2}{*}{Prompt-Tuning} & & 0.177 & 0.494 \\\cline{2-2}
& GPT-J-6B & \textbf{0.213} & \textbf{0.595} \\\hline

\end{tabular}
\caption{Results of automatic evaluation by distinct-1, 2. The prompt-tuned GPT-J-6B model generates the most diverse responses. ``Added'' and ``none'' mean whether the persona sentences are added to the input sentence or not.}
\label{tbl:distinct_auto_eval}
\end{table}

\begin{table*}[t]
\centering
\begin{tabular}{c|c|c||c|c|c}
\hline
\textbf{Eval Dataset} & \textbf{Training Method} & \textbf{Model} & \textbf{Fluency} & \textbf{Engagingness} & \textbf{Relevance}\\\hline
\hline 

\multirow{3}{*}{Persona Eval} & Fine-Tuning (none) & \multirow{2}{*}{GPT2-XL}  & 
3.52 (1.26) & 3.70 (1.22) & 3.30 (1.27)\\\cline{2-2}

& \multirow{2}{*}{Prompt-Tuning} & & 
3.82 (1.06) & 3.74 (1.17) & 3.62 (1.02)\\\cline{3-3}

& & GPT-J-6B & 
\textbf{3.90} (0.90) & \textbf{3.98} (0.95) & \textbf{3.82} (0.96)\\\hline

\multirow{3}{*}{General Eval} & Fine-Tuning (none) & \multirow{2}{*}{GPT2-XL} & 
3.93 (1.19) & \textbf{3.82} (1.20) & 3.77 (1.16)\\\cline{2-2}

& \multirow{2}{*}{Prompt-Tuning} & & 
\textbf{4.04} (1.01) & 3.81 (1.19) & \textbf{3.96} (1.13)\\\cline{3-3}

& & GPT-J-6B & 
3.98 (1.03) & 3.80 (1.01) & 3.89 (1.05)\\\hline
\multicolumn{3}{c||}{Human} & 4.31 (1.07) & 4.25 (1.06) & 4.36 (0.92)\\\hline
\end{tabular}
\caption{We evaluated the generated responses on a 5-point scale for fluency, engagingness, and relevance. We asked five workers to answer each question, and the averages of all answers and standard deviations (in parentheses) are shown. The prompt-tuned GPT-J-6B model scored highest in all aspects in the persona eval dataset. No significant differences were found in the general eval dataset.}
\label{tbl:human_eval}
\end{table*}

\begin{table*}[t]
\centering
\begin{tabular}{c|c|c||c|c|c|c}
\hline
\textbf{Eval Dataset} & \textbf{Training Method} & \textbf{Model} & \textbf{[1,2)} & \textbf{[2,3)} & \textbf{[3,4)} & \textbf{[4,5]} \\\hline
\hline 

\multirow{3}{*}{Persona Eval} & Fine-Tuning (none) & \multirow{2}{*}{GPT2-XL}  & 
0 & 5 & 33 & 12\\\cline{2-2}

& \multirow{2}{*}{Prompt-Tuning} & & 
0 & 7 & 41 & 2\\\cline{3-3}

& & GPT-J-6B & 
0 & 2 & 29 & 19\\\hline

\multirow{3}{*}{General Eval} & Fine-Tuning (none) & \multirow{2}{*}{GPT2-XL} & 
0 & 11 & 105 & 34\\\cline{2-2}

& \multirow{2}{*}{Prompt-Tuning} & & 
0 & 8 & 75 & 67\\\cline{3-3}

& & GPT-J-6B & 
0 & 1 & 91 & 58\\\hline
\end{tabular}
\caption{The generated responses were rated on a 5-point scale for persona consideration, and their distribution is shown. 1 is inconsistent with the persona, 3 is irrelevant to the persona, and 5 is in line with the persona. $[1,2)$ means the number of sentences scored between 1 and 2, including 1. In each setting, the number of samples from the persona eval dataset is 50 and that from the general eval dataset is 150.}
\label{tbl:human_eval_persona}
\end{table*}

\begin{table*}[t]
\centering
\small
\begin{tabular}{l|l}
\hline
\textbf{Utterance} & \textbf{Generated Response}\\\hline
\hline 

\begin{tabular}{l}are you tired ?\end{tabular} & \begin{tabular}{l}i am a little tired.\end{tabular}\\\hline

\begin{tabular}{l}cool , i am eating donuts . what do you do ?\end{tabular} & \begin{tabular}{l}i am retired, i am a gym teacher. i used to teach gym.\end{tabular}\\\hline

\begin{tabular}{l}how about the korean restaurant next to it ?\end{tabular} & \begin{tabular}{l}i heard it's very good.\end{tabular}\\\hline

\begin{tabular}{l}i used to cook paella for a kitchen in spain .\end{tabular} & \begin{tabular}{l}i love paella. that sounds great. \\ i bet it is a hard job to cook it right?\end{tabular}\\\hline

\begin{tabular}{l}the tv news is reporting a bank robbery .\end{tabular} & \begin{tabular}{l}i heard it on the radio. what a shock.\end{tabular}\\\hline

\begin{tabular}{l}what do you like to do in your spare time ?\end{tabular} & \begin{tabular}{l}i volunteer at a local soup kitchen, \\ helping people in need.\end{tabular}
\\\hline

\begin{tabular}{l}hi , i like to keep fit and work out 5 times a week\end{tabular} & \begin{tabular}{l}i like to eat cheeseburgers \\ and watch war documentaries\end{tabular} \\\hline

\begin{tabular}{l}stop producing cigarettes .\end{tabular} & \begin{tabular}{l}i like the idea, but how can we do that?\end{tabular}\\\hline

\end{tabular}
\caption{Response examples generated by the prompt-tuned GPT-J-6B model, which got the highest score in the manual evaluation. We can see that both of responses based on the persona (e.g., the sixth response) and responses unrelated to the persona (e.g., the first one) are generated.}
\label{tbl:generate_example}
\end{table*}

\begin{table}[t]
\centering
\small
\begin{tabular}{l}
\hline
\textbf{Persona Sentence}\\\hline
\hline 

i am a retired gym teacher.\\
i volunteer at a soup kitchen.\\
i was poor growing up.\\
cheeseburgers are my favorite food.\\
i like watching war documentaries.\\\hline

\end{tabular}
\caption{The persona used in the generated response examples in Table~\ref{tbl:generate_example}.}
\label{tbl:generate_example_persona}
\end{table}

\begin{table}[t]
\centering
\small
\begin{tabular}{c|c||c|c}
\hline
\textbf{Training Method} & \textbf{Model} & \textbf{Dist-1} & \textbf{Dist-2}\\\hline
\hline 

Fine-Tuning (none) & \multirow{2}{*}{GPT2-XL} & \textbf{0.134} & \textbf{0.379} \\\cline{1-1}
\multirow{2}{*}{Prompt-Tuning} & & 0.118 & 0.336 \\\cline{2-2}
& HyperCLOVA & 0.106 & 0.322 \\\hline

\end{tabular}
\caption{Results of automatic evaluation by distinct-1, 2 in experiments in Japanese. }
\label{tbl:distinct_auto_eval_ja}
\end{table}

\begin{table*}[t]
\centering
\begin{tabular}{c|c|c||c|c|c}
\hline
\textbf{Eval Dataset} & \textbf{Training Method} & \textbf{Model} & \textbf{Fluency} & \textbf{Engagingness} & \textbf{Relevance}\\\hline
\hline 

\multirow{3}{*}{Persona Eval} & Fine-Tuning (none) & \multirow{2}{*}{GPT2-XL}  & 
3.81 (1.12) & 3.63 (1.00) & 3.81 (1.06) \\\cline{2-2}

& \multirow{2}{*}{Prompt-Tuning} & & 
3.68 (1.23) & 3.67 (1.13) & 3.71 (1.17) \\\cline{3-3}

& & HyperCLOVA & 
\textbf{3.87} (1.11) & \textbf{3.92} (0.98) & \textbf{3.90} (1.08) \\\hline

\multirow{3}{*}{General Eval} & Fine-Tuning (none) & \multirow{2}{*}{GPT2-XL} & 
4.01 (0.96) & 3.82 (0.89) & 3.82 (1.00) \\\cline{2-2}

& \multirow{2}{*}{Prompt-Tuning} & & 
3.99 (1.09) & 3.68 (1.03) & 3.92 (1.08) \\\cline{3-3}

& & HyperCLOVA & \textbf{4.07} (1.01) & \textbf{3.86} (0.95) & \textbf{4.06} (0.99) \\\hline
\multicolumn{3}{c||}{Human} & 4.31 (1.07) & 4.25 (1.06) & 4.36 (0.92) \\\hline
\end{tabular}
\caption{Results of manual evaluation of fluency, engagingness, and relevance for the generated responses in the Japanese experiments. We asked five workers to answer each question, and the averages of all answers and standard deviations (in parentheses) are shown. Prompt-tuned HyperCLOVA scored highest in all aspects on both datasets.}
\label{tbl:human_eval_ja}
\end{table*}

\begin{table*}[t]
\centering
\begin{tabular}{c|c|c||c|c|c|c}
\hline
\textbf{Eval Dataset} & \textbf{Training Method} & \textbf{Model} & \textbf{[1,2)} & \textbf{[2,3)} & \textbf{[3,4)} & \textbf{[4,5]} \\\hline
\hline 

\multirow{3}{*}{Persona Eval} & Fine-Tuning (none) & \multirow{2}{*}{GPT2-XL}  & 
0 & 5 & 105 & 40\\\cline{2-2}

& \multirow{2}{*}{Prompt-Tuning} & & 
1 & 14 & 84 & 51\\\cline{3-3}

& & HyperCLOVA & 
0 & 18 & 77 & 55\\\hline

\multirow{3}{*}{General Eval} & Fine-Tuning (none) & \multirow{2}{*}{GPT2-XL} & 
0 & 8 & 122 & 20\\\cline{2-2}

& \multirow{2}{*}{Prompt-Tuning} & & 
0 & 14 & 115 & 21\\\cline{3-3}

& & HyperCLOVA & 
0 & 19 & 125 & 6\\\hline
\end{tabular}
\caption{Distribution of manually evaluated persona consideration in Japanese. In each setting, the number of samples is 150 for both persona eval and general eval datasets.}
\label{tbl:human_eval_persona_ja}
\end{table*}

\subsection{Model Setup}
\label{ssec:model_setup}
To compare our prompt-tuning model with fine-tuning, we use the datasets in Section~\ref{ssec:dataset_creation} and tune the pre-trained models of GPT series. We use two model sizes: GPT2-XL (1.5B parameters) and GPT-J-6B~\citep{gpt-j}. Fine-tuning of the GPT-J-6B model is not tested due to the lack of GPU memory.

The hyperparameters for prompt-tuning are based on the settings of ~\citep{lester-etal-2021-power}. 
The length of the persona info tokens was set to 200 based on the results of preliminary experiments.
The strategy for generating the response sentences is the greedy search.
The number of epochs was set to a value such that the loss during learning converges.
For fine-tuning, we experimented with two methods: one is to input only dialogue pairs, and the other is to add persona sentences before the dialogue pair's utterance and then input it into the model.
Other hyperparameter values are given in Appendix~\ref{sec:appendix_hyperparameter}.


\subsection{Results}
We input the utterances of dialogue pairs from the evaluation datasets into the trained models. We automatically evaluate the diversity of the generated responses and manually assess whether the responses
are natural and based on the persona. 

\subsubsection{Automatic Evaluation}
\label{ssec:auto_eval}
We evaluate the diversity of the generated responses by distinct-N~\citep{li-etal-2016-diversity}.
The values of distinct-1 and distinct-2 are shown in Table~\ref{tbl:distinct_auto_eval}.
The evaluation values are the average of all the generation results of the persona, general eval datasets from each model corresponding to the three types of personas. The results show that the GPT-J-6B model trained by prompt-tuning generates the most diverse responses. 
In fine-tuning, we also find that the results are better when persona sentences are not added to the input, similar to the experimental results using the seq2seq model in the experiments by ~\citet{zhang-etal-2018-personalizing}.




\subsubsection{Manual Evaluation}
We use Amazon Mechanical Turk to manually evaluate whether the generated responses are natural and persona-based.
Following the method of ~\citet{zhang-etal-2018-personalizing}, the responses are rated on a 5-point scale on four aspects: fluency, engagingness, relevance, and persona consideration.
We ask five workers to answer each question.
In each setting, the number of samples from the persona eval dataset is 50 and that from the general eval dataset is 150.
An example of tasks given to workers is shown in Appendix~\ref{sec:appendix_crowdsourcing}.

The results of the first three aspects are shown in Table~\ref{tbl:human_eval}.
The human scores are taken from the experiments by ~\citet{zhang-etal-2018-personalizing}.
In fine-tuning, when persona sentences are added to the input, the automatic evaluation results are worse than when they are not, and thus we only evaluate the models without persona sentences.
From Table~\ref{tbl:human_eval}, the manual evaluation results using the persona eval dataset show that the prompt-tuned GPT-J-6B model achieves the best scores in all aspects.
This can be attributed to the fact that the larger the model size was, the more knowledge was stored in the model through pre-training, and the more natural responses were generated by using this knowledge. 
Manual evaluation using the general eval dataset did not produce a significant difference.
This is probably because most of the utterances in the general eval dataset are short and general, such as greetings, and the responses are also short and simple sentences.

For persona consideration, the distribution of the evaluation results is shown in Table~\ref{tbl:human_eval_persona}, where 1 is inconsistent with the persona, 3 is irrelevant to the persona, and 5 is in line with the persona.
The average score of persona consideration is calculated for each generated response.
Table~\ref{tbl:human_eval_persona} shows that the majority of the generated responses are based on personas.

Comparing fine-tuning and prompt-tuning with the largest models that can be trained with a fixed GPU memory size, we can say that prompt-tuning can build a dialogue system with more natural responses based on the persona.

Table~\ref{tbl:generate_example} shows response examples generated by the prompt-tuned GPT-J-6B model, which got the highest score in the manual evaluation. These responses are generated from the model trained with the dialogue pairs based on persona sentences shown in Table~\ref{tbl:generate_example_persona}.
We can see that training with small training datasets of only a few hundred pairs can produce a response with a natural and consistent personality, as shown in Table~\ref{tbl:generate_example}.

\subsection{Experiments in Japanese}
For our Japanese experiments, we use two datasets: JPersonaChat and JEmpatheticDialogues~\citep{sugiyama2021empirical}.\footnote{\url{https://github.com/nttcslab/japanese-dialog-transformers}}
As in the English experiments, three personas are used, and the number of dialogue pairs from JPersonaChat are 527, 525 and 525, respectively.
To create training datasets, the same process as in the English experiments is used. Since most of the utterances in JEmpatheticDialogues are shorter and more general than those in JPersonaChat, we did not set any conditions for adding the utterances from JEmpatheticDialogues to the training datasets.
The ratio of the training datasets is set to 1:10 based on the results of preliminary experiments.
For the models, we use GPT2-XL\footnote{\url{https://huggingface.co/rinna/japanese-gpt-1b}} with 1.3B parameters and HyperCLOVA~\citep{kim-etal-2021-changes}, a GPT3-like model with 6.9B parameters.

In the automatic evaluation results shown in Table~\ref{tbl:distinct_auto_eval_ja}, in contrast to the English experiments, HyperCLOVA, which has a higher number of parameters, tends to score lower. This can be attributed to the fact that there were many instances in which HyperCLOVA begins its response with back-channeling.

Table~\ref{tbl:human_eval_ja} shows the average scores for the three aspects within the manual evaluation results.
For both the persona eval dataset and general eval dataset, the HyperCLOVA model with prompt-tuning scored the highest.
The distribution of persona consideration is shown in Table~\ref{tbl:human_eval_persona_ja}. As in the English experiments, many responses are based on the persona and few are inconsistent with the persona.
Generated response examples are shown in Appendix~\ref{sec:appendix_generated_sample_Ja}.

%% file: conclusion.tex
\section{Conclusion}
\label{sec:conclusion}
We proposed a method for prompt-tuning a pre-trained language model using dialogue data based on a single persona.
Automatic and manual evaluations showed that we could construct a dialogue system that can respond more naturally and persona-based, with less computational resources than fine-tuning. 
Compared to the generated responses in English, those in Japanese look natural due to the larger persona dataset.
In the future, this method can be used not only to add personality to a dialogue system but also to build a dialogue system to generate responses with emotions by making a prompt for each emotion.

%% file: acknowledgements.tex
\section*{Acknowledgements}
This work was supported by a joint research grant from LINE Corporation.

%% file: appendix.tex
\appendix
\onecolumn
\section{Generated response examples in Japanese}
\label{sec:appendix_generated_sample_Ja}
Response examples that generated by prompt-tuned HyperCLOVA are shown in Table~\ref{tbl:generate_example_ja}. 
These responses are generated from the model trained with the dialogue pairs based on persona sentences shown in Table~\ref{tbl:generate_example_persona_ja}.

\begin{table*}[!h]
\centering
\small
\begin{tabular}{l|l}
\hline
\textbf{Utterance} & \textbf{Generated Response}\\\hline
\hline 

\begin{tabular}{l}\begin{CJK}{UTF8}{ipxm}こんにちは。最近パーマをかけて\end{CJK}\\\begin{CJK}{UTF8}{ipxm}イメチェンしたから今度みてほしいな。\end{CJK}\\(Hello. I recently got a perm and a makeover,\\ so I hope you'll take a look at it sometime.)\end{tabular} &
\begin{tabular}{l}\begin{CJK}{UTF8}{ipxm}いいよ。いつがいい?わたしは家政婦の仕事が\end{CJK}\\\begin{CJK}{UTF8}{ipxm}あるから、夕方以降なら大丈夫だよ。\end{CJK}\\(Okay. When is a good time? I have to work\\as a housekeeper, so I can come after the evening.)\end{tabular}\\\hline

\begin{tabular}{l}\begin{CJK}{UTF8}{ipxm}こんにちは!ウサギ飼ってるんですね、羨ましい。\end{CJK}\\(Hello! I see you have a rabbit, I'm jealous.)\end{tabular} & 
\begin{tabular}{l}\begin{CJK}{UTF8}{ipxm}はい、飼ってます。ペットショップで\end{CJK}\\\begin{CJK}{UTF8}{ipxm}一目惚れして、そのまま連れて帰りました。\end{CJK}\\(Yes, I have one. I fell in love with her\\ at first sight at a pet store and took her home with me.)\end{tabular}\\\hline

\begin{tabular}{l}\begin{CJK}{UTF8}{ipxm}絵の勉強をヨーロッパでやるとしたら、\end{CJK}\\\begin{CJK}{UTF8}{ipxm}どの国がいいんですか?\end{CJK}\\(If you were to study painting in Europe,\\ which country would you prefer?)\end{tabular} &
\begin{tabular}{l}\begin{CJK}{UTF8}{ipxm}ゴッホが好きなら、\end{CJK}\\\begin{CJK}{UTF8}{ipxm}オランダがいいんじゃないですか?\end{CJK}\\(If you like Van Gogh, why not Holland?)\end{tabular}\\\hline

\begin{tabular}{l}\begin{CJK}{UTF8}{ipxm}大丈夫、ありがとう。短かったんだけど、\end{CJK}\\\begin{CJK}{UTF8}{ipxm}突然真っ暗になったから本当に怖い思いをした。\end{CJK}\\(No problem, thanks. It was short, but I was really\\ scared because all of a sudden it went dark.)\end{tabular} &
\begin{tabular}{l}\begin{CJK}{UTF8}{ipxm}それは怖いね。無事でよかった。\end{CJK}\\\begin{CJK}{UTF8}{ipxm}それにしても、停電って本当に困るよね。\end{CJK}\\(That's scary. I'm glad you're safe. \\Still, power outages are really annoying, aren't they?)\end{tabular}\\\hline

\begin{tabular}{l}\begin{CJK}{UTF8}{ipxm}パイン飴ってそんなに昔からあるの?\end{CJK}\\(Has candy pineapple been around that long?)\end{tabular} &
\begin{tabular}{l}\begin{CJK}{UTF8}{ipxm}うん、子どもの頃から食べてたよ。\end{CJK}\\(Yeah, I've been eating it since I was a kid.)\end{tabular}\\\hline

\end{tabular}
\caption{Response examples generated by prompt-tuned HyperCLOVA.}
\label{tbl:generate_example_ja}
\end{table*}

\begin{table}[!h]
\centering
\small
\begin{tabular}{l}
\hline
\textbf{Persona Sentence}\\\hline
\hline 
\begin{CJK}{UTF8}{ipxm}私は家政婦です。\end{CJK}
(I am a housekeeper.)\\

\begin{CJK}{UTF8}{ipxm}私は姉と暮らしています。\end{CJK}
(I live with my sister.)\\

\begin{CJK}{UTF8}{ipxm}私のペットはウサギです。\end{CJK}
(My pet is a rabbit.)\\

\begin{CJK}{UTF8}{ipxm}私が尊敬する人は、画家のゴッホです。\end{CJK}
(The person I admire is the painter Van Gogh.)\\

\begin{CJK}{UTF8}{ipxm}私は美術部に入っていました。\end{CJK}
(I was in the art club.)\\\hline

\end{tabular}
\caption{The persona used in the generated response examples in Table~\ref{tbl:generate_example_ja}.}
\label{tbl:generate_example_persona_ja}
\end{table}

\section{Hyperparameter}
\label{sec:appendix_hyperparameter}
Table~\ref{tbl:hyperparameter} shows hyperparameters during model training in our experiment.

\begin{table}[h]
\centering
\small
\begin{tabular}{l|c|c|c|c}
\hline
\textbf{Hyperparameter} & \textbf{Fine-Tuning (En)} & \textbf{Prompt-Tuning (En)} & \textbf{Fine-Tuning (Ja)} & \textbf{Prompt-Tuning (Ja)}\\\hline
\hline 
Optimizer & Adam & Adam & Adam & Adam \\\hline
Learning Rate & 5e-5 & 1e-3 & 1e-5 & 1e-3 \\\hline
\end{tabular}
\caption{Hyperparameters during model training in our experiment.}
\label{tbl:hyperparameter}
\end{table}

\clearpage
\section{An example of tasks used in crowdsourcing}
\label{sec:appendix_crowdsourcing}
\begin{figure}[h]
    \centering
    \includegraphics[width=1\linewidth]{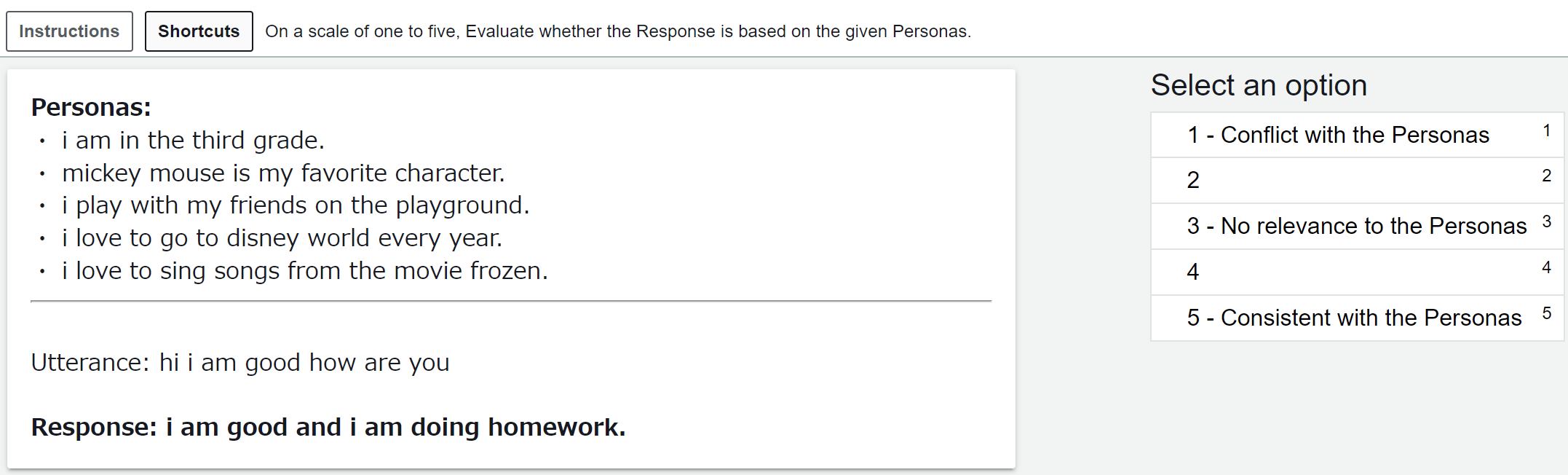}
    \caption{An example of tasks given to workers on Amazon Mechanical Turk for the manual evaluation.}
    \label{fig:crowdsourcing}
\end{figure}

Figure~\ref{fig:crowdsourcing} shows an example of tasks used in crowdsourcing.

